\documentclass[10pt,twocolumn,letterpaper]{article}

\usepackage[T1]{fontenc}
\usepackage{CJKutf8}
\usepackage[english]{babel}

\usepackage{iccv,times,graphicx,tabulary,multirow,subfig,xspace,xcolor}
\usepackage{amsthm,amssymb,fixmath,mathtools,nicefrac}
\usepackage[pagebackref=true,breaklinks=true,letterpaper=true,colorlinks,
  citecolor=citecolor,bookmarks=false]{hyperref}
\usepackage[normalem]{ulem} 
\definecolor{citecolor}{RGB}{34,139,34}

\begin{document}
\begin{CJK}{UTF8}{mj}
\title{HintPose}
\track{Keypoint Detection}

\author{Sanghoon Hong \qquad Hunchul Park\\
Kakao Brain\\
Seongnam, Gyeonggi, South Korea\\
{\tt\small \{sanghoon.hong, robert.p\}@kakaobrain.com}
\and
Jonghyuk Park \qquad Sukhyun Cho \qquad Heewoong Park \\
Seoul National University\\
Seoul, South Korea\\
{\tt\small \{chico2121, chosh90, hee188\}@snu.ac.kr}
}

\maketitle

\begin{abstract}

Most of the top-down pose estimation models assume that there exists only one person in a bounding box. However, the assumption is not always correct. In this technical report, we introduce two ideas, instance cue and recurrent refinement, to an existing pose estimator so that the model is able to handle detection boxes with multiple persons properly. When we evaluated our model on the COCO17 keypoints dataset, it showed non-negligible improvement compared to its baseline model. Our model achieved 76.2 mAP as a single model and 77.3 mAP as an ensemble on the test-dev set without additional training data. After additional post-processing with a separate refinement network, our final predictions achieved 77.8 mAP on the COCO test-dev set.

\end{abstract}
 
\section{Introduction}

\begin{figure}
    \centering
    \includegraphics[scale=0.25]{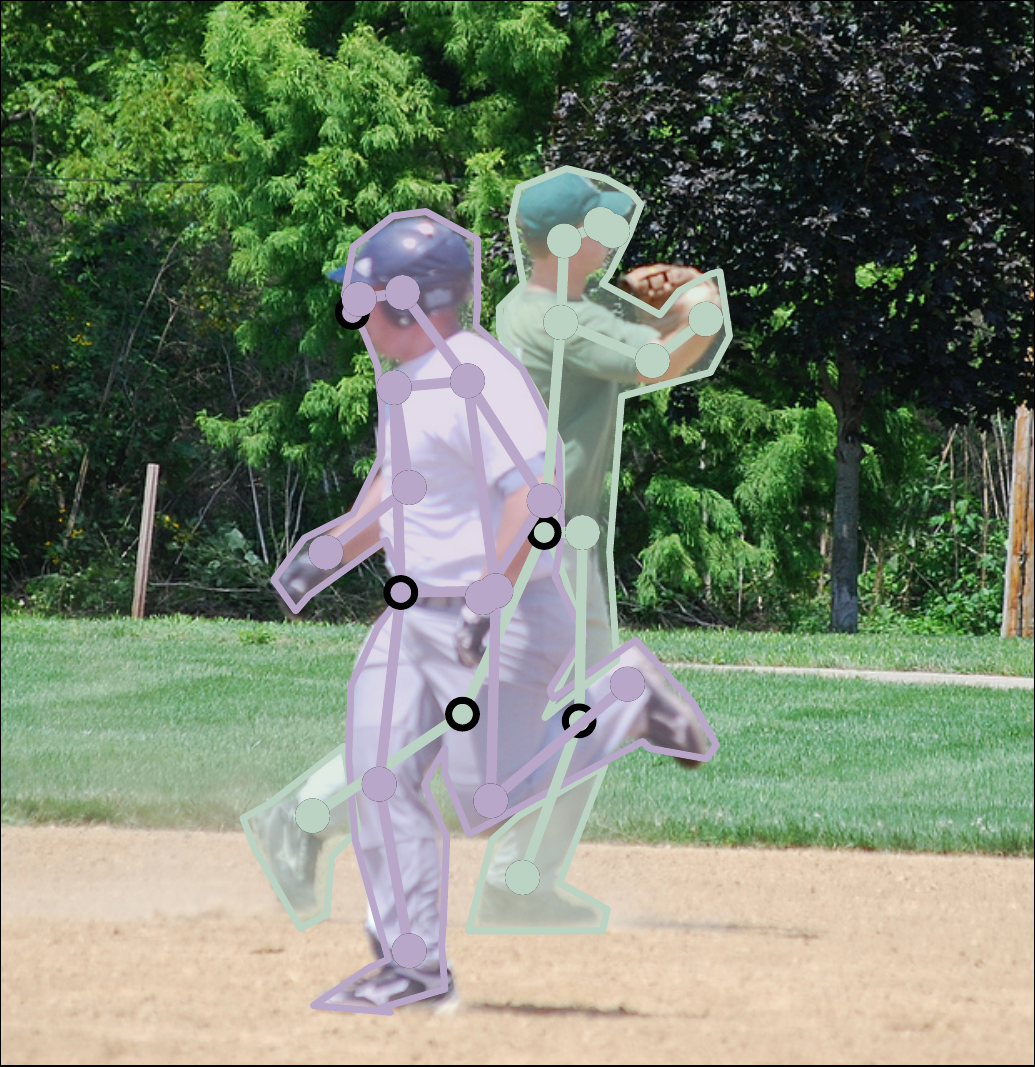}
    \includegraphics[scale=0.25]{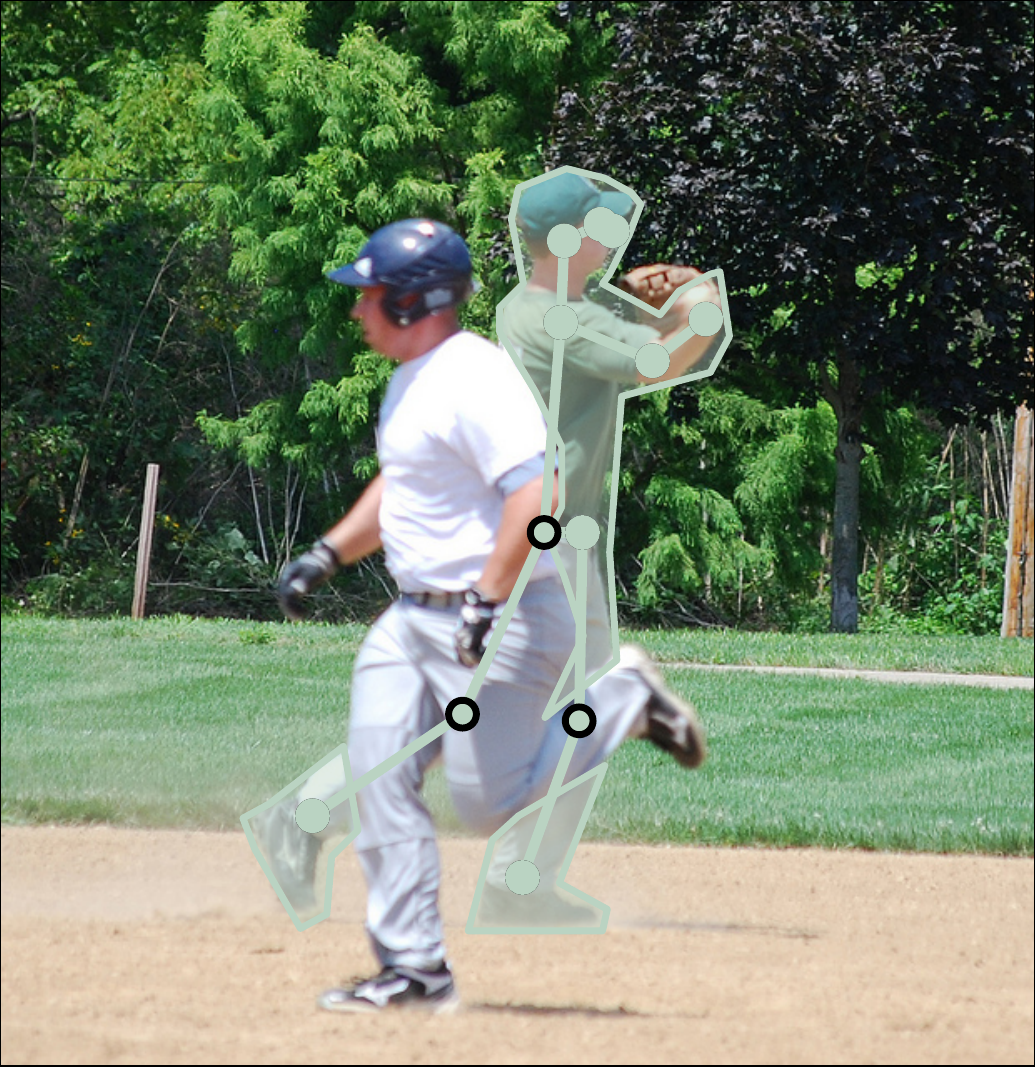}
    \includegraphics[scale=0.25]{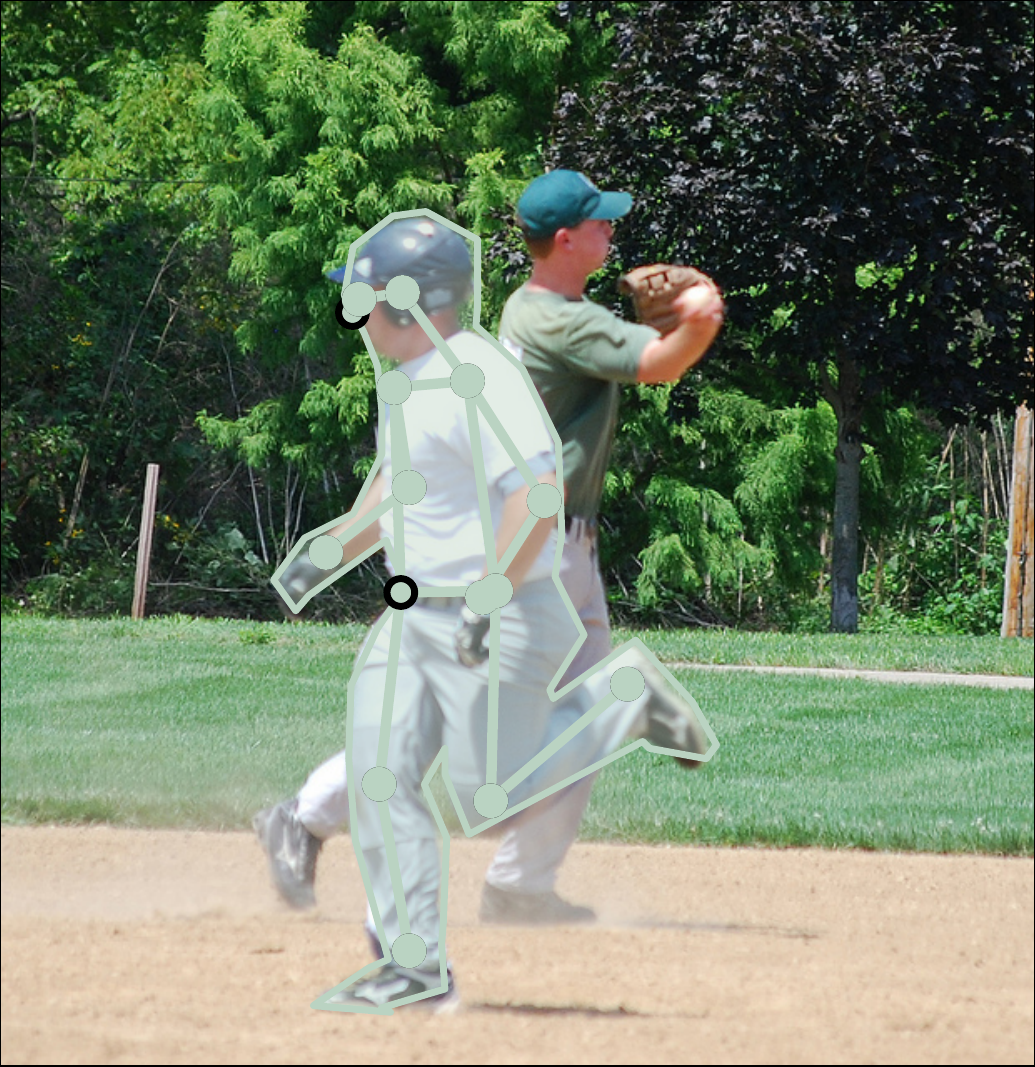}
    \caption{An example of two overlapping persons in one bounding box.}
    \label{fig:inst_cue}
\end{figure}

Most of the top-down pose estimation models, such as HRNet\cite{sun2019deep}, CPN\cite{chen2018cascaded}, Mask R-CNN\cite{he2017mask} generate predictions assuming there exists only one person in one person detection box. However, multiple persons can reside in a box in some cases, as shown in Fig \ref{fig:inst_cue}, and it is difficult to say which one is a dominant target person. It becomes more difficult as a model processes cropped images from expanded detection boxes to capture context information. This phenomenon makes single-person pose estimation as an ill-posed problem, in which there exist multiple solutions.

To alleviate this issue, we introduce two ideas; the first idea is to add \textit{instance cue} on input which specifies a target person in a box, and the other is to design a recurrent network so that a model can refine its predictions using the outputs from previous hops as a hint for a target person.

\section{HintPose}

\begin{figure*}
    \centering
    \includegraphics[width=\textwidth]{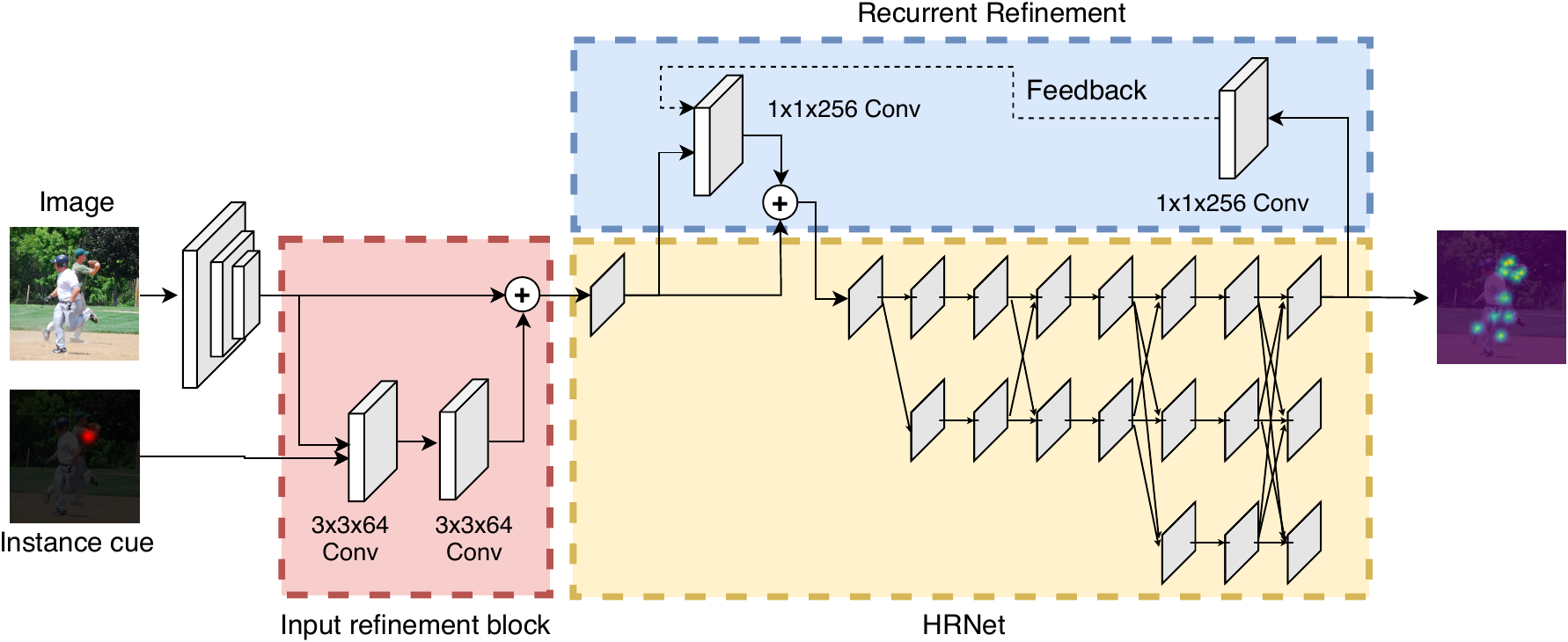}
    \caption{HintPose Network Architecture}
    \label{fig:architecture}
\end{figure*}

Figure \ref{fig:architecture} shows our network structure. We adopt HRNet\cite{sun2019deep} as a baseline architecture since, to the best of our knowledge, it is the state-of-the-art model with open-sourced codes. Then we add an input refinement block so that the model can handle an external instance cue for a target person. We also create a feedback connection from the output of the network to the intermediate feature map so that the model can refine its outputs using its previous predictions. For each of both modifications, we just add two simple convolutional layers with a residual connection to reuse ImageNet pre-trained models provided by the official HRNet respository\footnote{https://github.com/leoxiaobin/deep-high-resolution-net.pytorch}.

\subsection{Instance Cue}
As mentioned earlier, there may exist two or more persons in a bounding box from annotated data or a detection box from an object detector. To specify a target person explicitly, a cropped image and an instance cue embedding are fed into our network. The embedding is a single channel Gaussian heatmap which has a peak located on a target person.

The input refinement block in HintPose aggregates image features and instance cue embedding and updates feature maps with element-wise summation.

Instance cues can be derived from ground-truth keypoints or instance segmentation maps during training time. At inference time, they can be generated from the outputs of other instance segmentation or keypoints estimation models, or it is also possible to train another simple network to predict them directly.

\subsection{Recurrent Refinement}
In addition to providing an external instance cue, it is also possible to use the outputs of the model itself as a hint for a target person. We adopted the structure of Feedback Network\cite{li2019srfbn} and designed our model to have a recurrent connections so that it can refine its outputs using its previous predictions.

We added two $1 \times 1$ convolutional blocks onto the baseline network; one is to update feature maps using information from the previous output and another is to extract meaningful information for the next hop. We built the connection to feed back onto features after \texttt{layer1} of HRNet so that the improved features can be processed in all the different scales of HRNet and have smaller memory footprint.

\section{Experiments}
\subsection{Training \& Evaluation Details}
\paragraph{Training}
While training the network, we generated instance cues by randomly selecting a joint among ground-truth joints and augmenting its x, y position.
For a model with recurrent refinement, the model is evaluated for three hops, and all of its prediction outputs are used to compare with a ground-truth heatmap and to compute mean squared error.

We trained our models with the COCO training set only and used the same hyper-parameters provided by the official HRNet repository.

\paragraph{Evaluation}
We used MMDetection toolbox\cite{mmdetection} and Hybrid Task Cascade (HTC)\cite{chen2019hybrid} + HRNetV2p-W48 model to generate detection boxes on the COCO17 validation and test sets. Its detection accuracy is 47.0 mAP (60.5 mAP for `person' category) on the COCO17 validation set. We ignored bounding boxes smaller than $32 \times{} 32$.

To generate instance cues during an evaluation, we generated image-level joint heatmaps using MultiPoseNet\cite{kocabas18prn} and considered local peaks in each detected bounding box as instance cues. When there are multiple cues in a bounding box, the same cropped image are fed into the model multiple times with each cue.

For models with recurrent refinement, the output heatmaps after three hops are used to compute the final predictions.

Other hyper-parameters and post-processing, including person scoring and OKS-base NMS, are kept the same with the original HRNet.

\subsection{COCO Keypoints Detection}

We evaluated our models with the COCO\cite{lin2014coco} 2017 Keypoints validation and test-dev sets.

\paragraph{Validation set}

\begin{figure}
    \centering
    \includegraphics[scale=0.20]{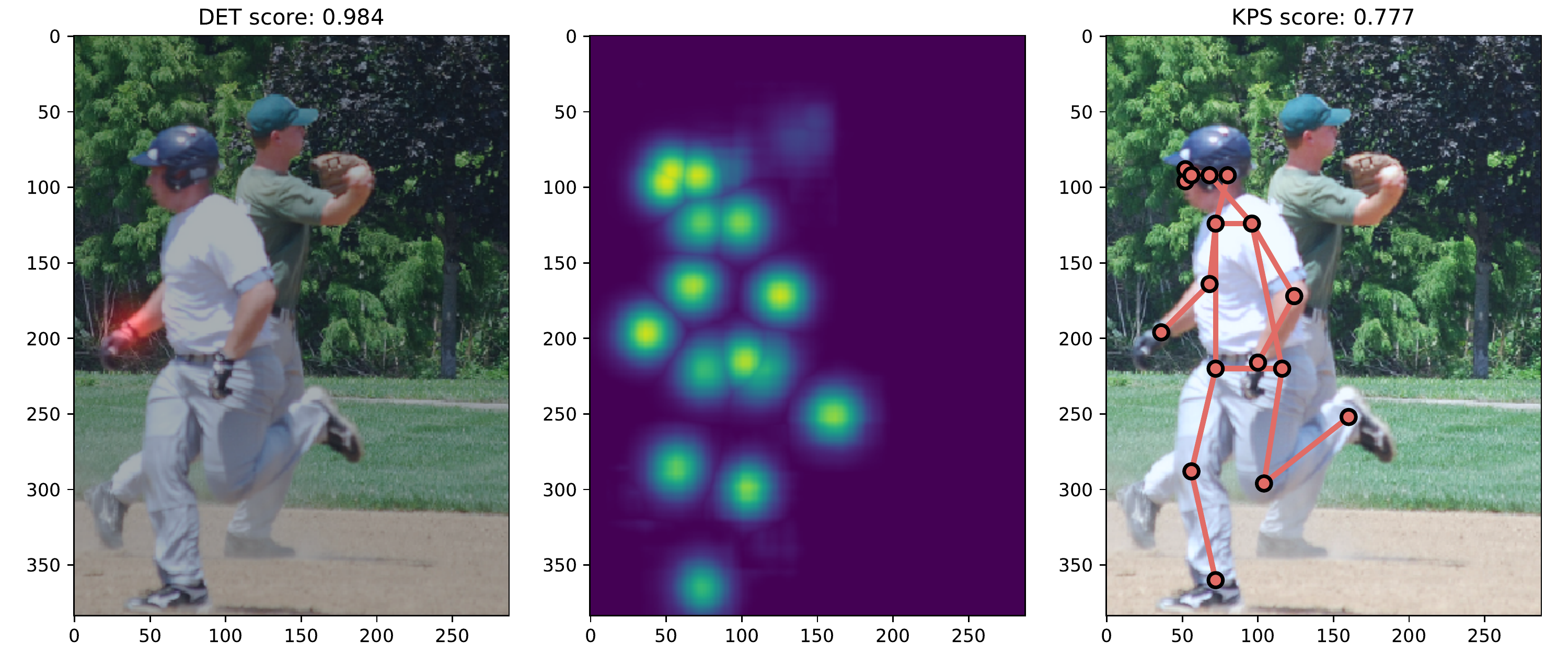}
    \includegraphics[scale=0.20]{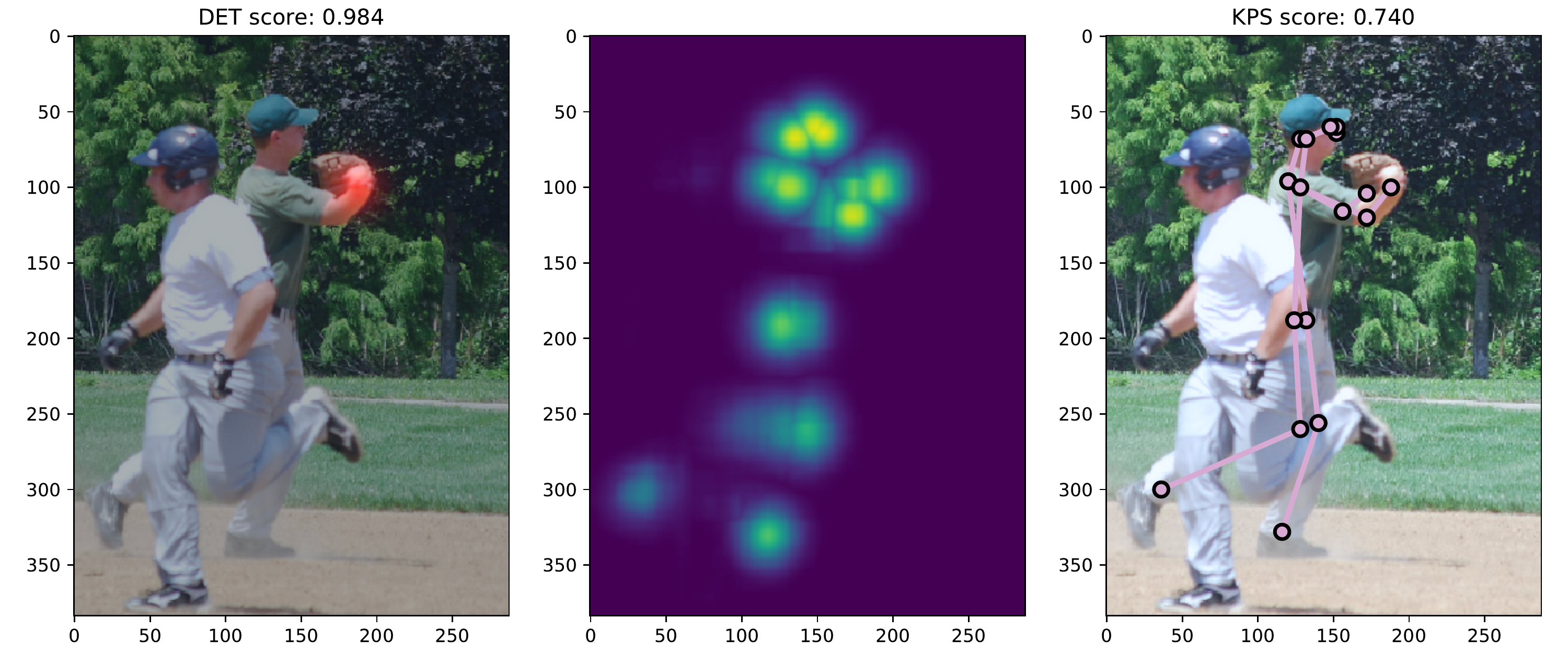}
    \caption{Two different predictions results in one person detection box. Images on the first column show cropped image and instance cue embedding. Images on the second and last column describes predicted heatmaps and skeletons.}
    \label{fig:inst_cue2}
\end{figure}

Figure \ref{fig:inst_cue2} shows two different predictions from the same input image with two different instance cues. The predictions of our model vary as instance cue moves from one person to another.

When they are evaluated on the COCO validation set, Both \textit{Instance Cue} and \textit{Recurrent Refinement} improved the performance of our pose estimation model. Moreover, the improvement increased as the modifications are applied together.

\begin{table}[]
    \centering
    \begin{tabular}{l|cc}
        Method & AP & AR \\
        \hline
        HRNet (Baseline) & 76.7 & 81.2 \\
        Instance cue & 77.1 & 81.5 \\
        Recurrent refinement & 78.0 & 82.3 \\
        I.C. + R.R. & \textbf{78.1} & \textbf{82.6} \\
    \end{tabular}
    \caption{COCO Validation results. `I.C. + R.R.' stands for a model with both instance cue and recurrent refinement.}
    \label{tab:coco_val}
\end{table}

\begin{table}[]
    \centering
    \begin{tabular}{l|cc}
        Method & AP & AR \\
        \hline
        HRNet (Baseline) & 75.4 & 80.2 \\
        Instance cue & 75.4 & 80.2 \\
        Recurrent refinement & 76.2 &  81.0 \\
        I.C. + R.R. & 76.2 & 81.0 \\
        \hline
        Ensemble model & 77.3 & 81.8 \\
        Ensemble model + PoseFix & \textbf{77.8} & \textbf{82.2} \\
    \end{tabular}
    \caption{COCO Test-dev results. `I.C. + R.R.' stands for a model with both instance cue and recurrent refinement.}
    \label{tab:coco_testdev}
\end{table}

\paragraph{Test-dev set}
Our model showed a significant improvement of +0.8 mAP compared to the baseline HRNet. Moreover, when we ensemble 6 different models\footnote{Averaged heatmaps of the followings are used for post-processing; two models with instance cue and recurrent refinement, two models with instance cue only and two models with recurrent refinement only.}, our models achieved 77.3 mAP. We also refined our predictions with PoseFix\cite{Moon_2019_CVPR_PoseFix} and the final predictions achieved 77.8 mAP on the COCO test-dev set.

One observation is that the improvements from Instance cue are not as meaningful as those on the COCO validation set. We hypothesized that the number of crowded bounding boxes is less significant on the test-dev set.

\section{Discussions \& Future Works}
In this technical report, we introduced two different methods to handle multiple, overlapped person instances in one bounding box. Our modifications can be applied to existing top-down pose estimation models by adding a couple of convolutional blocks. When we evaluated our model on the COCO keypoints dataset, we observed non-negligible performance improvements thanks to our HintPose method.

Our future work will include evaluating our model on other datasets and figuring out how our method performs when it is applied to other top-down models. It is known that crowded scenes are not dominant in the COCO Keypoint dataset\cite{li2018crowdpose}. Therefore, we expect that more significant improvements can be observed when our model is evaluated on a dataset with more occlusions such as CrowdPose\cite{li2018crowdpose}.

It is also possible to improve our model with a better learning strategy as \cite{li2019srfbn} showed that curriculum learning is vital to train its feedback network structure. Another way to improve our model can be to use a different type of instance cue, such as segmentation maps.

{\small \bibliographystyle{ieee_fullname} \bibliography{egbib}}

\begin{thebibliography}{10}\itemsep=-1pt

\bibitem{chen2019hybrid}
Kai Chen, Jiangmiao Pang, Jiaqi Wang, Yu Xiong, Xiaoxiao Li, Shuyang Sun,
  Wansen Feng, Ziwei Liu, Jianping Shi, Wanli Ouyang, Chen~Change Loy, and
  Dahua Lin.
\newblock Hybrid task cascade for instance segmentation.
\newblock In {\em The IEEE Conference on Computer Vision and Pattern
  Recognition (CVPR)}, 2019.

\bibitem{mmdetection}
Kai Chen, Jiaqi Wang, Jiangmiao Pang, Yuhang Cao, Yu Xiong, Xiaoxiao Li,
  Shuyang Sun, Wansen Feng, Ziwei Liu, Jiarui Xu, Zheng Zhang, Dazhi Cheng,
  Chenchen Zhu, Tianheng Cheng, Qijie Zhao, Buyu Li, Xin Lu, Rui Zhu, Yue Wu,
  Jifeng Dai, Jingdong Wang, Jianping Shi, Wanli Ouyang, Chen~Change Loy, and
  Dahua Lin.
\newblock {MMDetection}: Open mmlab detection toolbox and benchmark.
\newblock {\em arXiv preprint arXiv:1906.07155}, 2019.

\bibitem{chen2018cascaded}
Yilun Chen, Zhicheng Wang, Yuxiang Peng, Zhiqiang Zhang, Gang Yu, and Jian Sun.
\newblock Cascaded pyramid network for multi-person pose estimation.
\newblock In {\em The IEEE Conference on Computer Vision and Pattern
  Recognition (CVPR)}, 2018.

\bibitem{he2017mask}
Kaiming He, Georgia Gkioxari, Piotr Doll{\'a}r, and Ross Girshick.
\newblock Mask r-cnn.
\newblock In {\em The IEEE International Conference on Computer Vision (ICCV)},
  2017.

\bibitem{kocabas18prn}
Muhammed Kocabas, Salih Karagoz, and Emre Akbas.
\newblock Multi{P}ose{N}et: Fast multi-person pose estimation using pose
  residual network.
\newblock In {\em European Conference on Computer Vision (ECCV)}, 2018.

\bibitem{li2018crowdpose}
Jiefeng Li, Can Wang, Hao Zhu, Yihuan Mao, Hao-Shu Fang, and Cewu Lu.
\newblock Crowdpose: Efficient crowded scenes pose estimation and a new
  benchmark.
\newblock {\em arXiv preprint arXiv:1812.00324}, 2018.

\bibitem{li2019srfbn}
Zhen Li, Jinglei Yang, Zheng Liu, Xiaomin Yang, Gwanggil Jeon, and Wei Wu.
\newblock Feedback network for image super-resolution.
\newblock In {\em The IEEE Conference on Computer Vision and Pattern
  Recognition (CVPR)}, 2019.

\bibitem{lin2014coco}
Tsung-Yi Lin, Michael Maire, Serge Belongie, James Hays, Pietro Perona, Deva
  Ramanan, Piotr Doll{\'a}r, and C~Lawrence Zitnick.
\newblock Microsoft {COCO}: Common objects in context.
\newblock In {\em European Conference on Computer Vision (ECCV)}, 2014.

\bibitem{Moon_2019_CVPR_PoseFix}
Gyeongsik Moon, Juyong Chang, and Kyoung~Mu Lee.
\newblock Posefix: Model-agnostic general human pose refinement network.
\newblock In {\em The IEEE Conference on Computer Vision and Pattern
  Recognition (CVPR)}, 2019.

\bibitem{sun2019deep}
Ke Sun, Bin Xiao, Dong Liu, and Jingdong Wang.
\newblock Deep high-resolution representation learning for human pose
  estimation.
\newblock In {\em The IEEE Conference on Computer Vision and Pattern
  Recognition (CVPR)}, 2019.

\end{thebibliography}

\end{CJK}
\end{document}